# Using Large Pre-Trained Language Model to Assist FDA in Premarket Medical Device Classification

Zongzhe Xu
*Washington University in St. Louis*
St. Louis, USA
xuzongzhe@wustl.edu

**Abstract-** This paper proposes a possible method using natural language processing that might assist in the FDA medical device marketing process. Actual device descriptions are taken and matched with the device description in FDA Title 21 of CFR to determine their corresponding device type. Both pre-trained word embeddings such as FastText and large pre-trained sentence embedding models such as sentence transformers are evaluated on their accuracy in characterizing a piece of device description. An experiment is also done to test whether these models can identify the devices wrongly classified in the FDA database. The result shows that sentence transformer with T5 and MPNet and GPT-3 semantic search embedding show high accuracy in identifying the correct classification by narrowing down the correct label to be contained in the first 15 most likely results, as compared to 2585 types of device descriptions that must be manually searched through. On the other hand, all methods demonstrate high accuracy in identifying completely incorrectly labeled devices, but all fail to identify false device classifications that are wrong but closely related to the true label.

## I.  INTRODUCTION

As a major federal agency of the Department of Health and Human Services, Food and Drug Administration, also known as the FDA, has started to protect public health in the United States since the start of the last century[1]. One major component of the FDA's responsibilities is to regulate the marketing of new medical devices by ensuring the safety and efficacy of these premarket devices. Tests and evaluations were stringently done on these devices before they are allowed for transactions. However, as technology progresses, more and more healthcare experts and organizations start to adopt artificial intelligence in their process of research and development, and an increasing number of devices are created to handle a variety of tasks. In the Fiscal Year 2021, FDA reported receiving more than 18000 premarket submissions annually[2].

Due to the great variety and quantity of emerging medical devices, the difficulty and complexity of fulfilling the mission of the FDA escalate. The average review time of a 510(k)-submission increased from 100 days to 180 days from 2000 to 2014[3]. In response to this change, FDA started to turn to the assistance of machines in its process of regulation. According to Mason Marks[4], FDA is replacing physical laboratory and clinical experiments with computer model and simulation that depends on artificial intelligence, which greatly increases the efficacy of the reviewing process. However, there is still much debate around the legitimacy and applicability of utilizing such an automatic approach without human supervision.

On the other hand, artificial Intelligence is used in healthcare to perform more accurate diagnoses and improve the healthcare delivery process to patients. There have already been several cases in which computer diagnosis exceeds the performance of human experts such as spotting malignant tumors[5]. The increasing market of artificial intelligence in healthcare is further escalated by the Covid-pandemic, which brought public concerns for healthcare to a much higher altitude and a sheerly dropped amount in the workforce. Artificial intelligence in healthcare alone generates 6.9 billion USD in the year 2021 and is continuously growing with a compound growth rate of 46.2%[6]. The expected market size of this booming market is anticipated to reach 194.4 billion by the year 2030.

Like any fast-growing industry, AI in healthcare poses potential risks which require regulation. However, as stated by Simon Chesterman[7], such regulation is complicated by the pace of change, the wariness of constraining innovation, and the immense difference it will bring to the traditional regulatory models. Among all the changes that challenge the conventional regulation system, one of the most controversial is probably using AI for regulation. Algorithms are built in a way that eliminates unwanted human judgment variability in the regulation process. While they can also introduce discrimination independently from the dataset, they are trained on[8]. Another property that might be interesting to explore is whether this bias that the AI model exhibits is accumulative; that's it, if AI is used in AI regulation, will the discriminations transfer, or even worse, add up? Although many preliminary trials have been performed like the experiment of the "smart court" in Hangzhou, China in late 2019, such risks should always be wary of when conducting research in pertinent fields.

With these backgrounds, this publication will propose a new method that helps the FDA simplify its process of regulating the marketing of new medical devices. We will examine the possibility of applying deep learning networks, especially pre-trained word embedding and sentence embedding to the regulation process. The first section is going to be an introduction to relevant information. The second section will describe different sets of experiments and the applied methods, which include GPT-3 pre-trained embedding, FastText pre-trained embedding with subword information, and sentence embedding Siamese network. The third section will be an analysis of the experiment results. The final section will discuss the limitations of the methods used and ways to improve the results. Because this paper is almost solely based on pre-trained models and pre-trained embeddings, it also provides insight into how much we can exploit transfer learning and domain shifting in low-resource tasks.

## II. BACKGROUND

**Title 21 of Code of Federal Regulations (CFR), Parts 862-892:**

Title 21 of CFR is where FDA classifies and describes over 2500 distinct types of medical devices. For every kind of medical device, the CFR also provides a general description and the classification of this type of device, Class I or Class II, for example. Applicants need to find the matching type of their device along with which marketing pathway they need to be applying for from CFR, or they can submit a 513(g) request which is a service that the FDA provides that helps applicants to identify the regulation number of their devices that come with a user fee.

**Marketing Pathway:**

Premarket devices are classified into three classes according to risk, each corresponding to a different marketing pathway. Class I devices, which present the lowest risk, only require general controls most of the time. Class II devices require unique controls and a 510 (k) premarket notification submission, while Class III devices are required to go through premarket approval (PMA), the most stringent type of premarketing submission.

This paper proposes another way of simplifying the FDA premarket process by introducing automation at the beginning of the marketing pathway, device classification. Applicants must identify their device's class before submitting documents in these marketing pathways. However, the applicants need to place the class of their products according to the descriptions of different genres of devices from the classification panel of Title 21 of the Code of Federal Regulations. While there are over 2500 genres of medical devices on the classification panel, the only method an applicant can use to search the proper genre is keyword searching, which provides a somewhat inaccurate result. As a matter of fact, from the dataset pulled down from the FDA website, several records show erroneous classification of the device. While FDA can indeed assist in this classification process through methods like the 513(g) request, this process only adds more burden to the already overloading system.

In this paper, three methods are adopted to automatically classify a new medical device by matching its description directly to the reports of types of devices in Title 21 of the CFR, Parts 862-892. With this approach, applicants no longer need to manually search the CFR classification panel for the regulation number of their device. All it requires is to write a brief description for the device, and they can select the desired device type from the top few results that match most accurately with the description they provide. This will boost productivity by reducing the inaccuracy in the submitted marketing application and free FDA from the 513(g) request, which is relatively inefficient.

### III. EXPERIMENT SETUP

The task is to compare the device descriptions in the CFR and the device description provided by the applicant by computing their similarity. The target device description uses pre-trained models and embeddings to produce sentence embeddings. Similarity scores are yielded between the embeddings. The expectation is that the type of device in CFR that has a higher similarity score matches better the target device description. This paper sets two experiment goals. 1. Correctly identify the few most probable device types for the target device. 2. Identify devices that present misclassifications in the FDA 510(k) submissions.

**Dataset**

The dataset of this experiment consists of two parts. For the CFR device description dataset, all types of devices, 2585 in total, and their corresponding genres and descriptions are crawled from Title 21 of the CFR dataset on the FDA classification panel. Then, we randomly selected 30 devices from the FDA premarket application database in 2021 and manually recorded the appropriate device descriptions and their regulation numbers. The second part of the dataset comes from the 510(k) dataset, an easier accessible source than the premarket approval (PMA). Because the purpose of this experiment is to match two device descriptions, which don't vary from 510(k) to PMA, this dataset is considered representative. Moreover, from the randomly selected 510 (k) devices description, some data points are spotted with inaccurate regulation numbers because their described device types do not match their device descriptions. Such data points are labeled manually and analyzed further for the second experiment goal.

One major problem with the 510(k) submission is that the format is not fixed. The descriptions of devices might appear anywhere in the submission documents. Therefore, it is hard to design an automatic method that extracts the desired information from these documents, so the volume of the second dataset is limited. Since we are performing a text classification with a huge number of output classes, with over 2500 labels, such capacity of the dataset is far from enough to perform supervised learning with a super large SoftMax layer.

Another problem that needs consideration is that the device descriptions are highly restricted. Almost all descriptions contain unique words like the name of the device and professional biology terms like "nephroma." Such restrictedness is not covered by our pre-trained embeddings and might cause inaccuracies during the similarity comparison.

Regarding this issue, pre-trained models and pretrained embedding are the main methods used in this paper. However, this paper also tries to train a FastText supervised model initialized with FastText pre-trained embeddings on the descriptions from the CFR device description dataset, which includes one example for each device type. The purpose is to compare and test whether a supervised model performs better than pretrained unsupervised methods using the same information.

**Metrics**

After similarity scores are obtained between each device type in the CFR and the target device, the similarity rank of the original label of the target device will be recorded. For example, suppose a device is labeled as "Abnormal hemoglobin assay" in the 510 (k) submission. In that case, similarity sorting is performed and returns a list of the possible device type. The similarity of "Abnormal hemoglobin assay" is the fifth highest. Then the device is labeled 5. Therefore, the lower the label number, the better the algorithm matches the original label in the 510(k) submissions. The rankings of the 30 510(k) dataset are outputted in such an approach for analysis purposes. In this case, the baseline value for each device rank is half of the total number of the categories in the CFR, 1297.5, which is the average rank for random guessing.

To determine inaccurately labeled data, the author manually compares the device's original description in the submission and the description of the tagged device type in the CFR. Any comparison that looks inaccurate or irrelevant will be marked. This can be considered as a human recognition level baseline.

**Method**

| | attention | word level/sentence level | Tokenization/Can handle OOV words | model type | model constitution |
|---|---|---|---|---|---|
| fasttext word embedding | No | word level | space separated/No | skipgram | \ |
| GPT-3 similarity embedding | Yes | sentence level | BPE/No | autoregressive | transformer decoder |
| GPT-3 semantic search embedding | Yes | sentence level | BPE/No | autoregressive | transformer decoder |
| Sentence BERT | Yes | sentence level | WordPiece/No | masked tokens | transformer encoder |
| Sentence RoBERTa | Yes | sentence level | Byte level BPE/Yes | masked tokens | transformer encoder |
| Sentence T5 | Yes | sentence level | SentencePiece+Unigram/No | masked text filling | encoder-decoder |
| Sentence MPNet | Yes | sentence level | WordPiece/No | masked+permuted | transformer decoder |

*Table 1: features and characteristics of models we tested in this paper.*

Pre-trained word embeddings have been proven to work well on text classifications[9]. It is word-to-vector mappings that create a vector space that match the semantic meaning of words. The model will be trained with faster speed and higher accuracy with this pre-learned knowledge. As an essential part of transfer learning, it has been used widely in NLP-related tasks. Ye Qi, et al[10] evaluated the effect of pre-trained word embeddings on neural machine translation in 2018. In the year 2019, Hayashi, Tomoki, et al[11, p.], on the other hand, used the pre-trained BERT model in text-to-speech synthesis, which generated a significantly more natural result than the baseline model. In the same year, Mingzhe Du, et al[12], evaluate the performance of different word embeddings (e.g., context-free and contextual embedding) on email intention detection. This paper will use FastText pretrained embedding with subword information.

Pre-trained models and dynamic word embedding, on the other hand, provide an improvement upon the representations of static pre-trained word embedding by incorporating context, documents, and language information using methods like autoencoder and auto-regression. Models like XLNet and ERNIE have been designed that show significantly higher performance on standard NLP tasks. In this paper, sentence transformers, a type of model that is built upon transformer-based pre-trained models with the Siamese network to derive sentence similarity, are chosen to perform our experiment. GPT-3 embedding is also used in the experiments to find similarities between device descriptions.

After observations of the dataset, many device descriptions contain a reference to other devices along with the regulation numbers of those devices. One such example is "…may be labeled for use with breathing circuits made of reservoir bags (§ 868.5320), oxygen cannulas (§ 868.5340) …" All contents containing such regulation numbers, along with other contents inside the parenthesis containing the regulation numbers, are removed from the dataset since they contribute no information to the model.

**FastText pre-trained embeddings**

FastText pre-trained embedding is developed in 2017 and is known for its capability of incorporating morphology (subword) information in its vector representations. It utilizes a skipgram model except that it represents the word embedding as the sum of several character n-grams during unsupervised training on the corpus, which contains the subword information[13]. This paper selects the FastText embeddings with subword information pre-trained on Common Crawl which has a dimension of 300.

When constructing the embeddings for the device descriptions, pre-trained embeddings are found for each word in the descriptions and the average of these word embeddings is taken with TF-IDF weighting to form the sentence embeddings for the description. TF-IDF is short for term frequency-inverse document frequency. It measures how important a word is in a sentence compared to the entire document. Unique words in each description are given a high value of weighting while common words across all documents like "is", "the", and "in" are given a small value so we can ignore their effects on our sentence embeddings. In this sense, the sentence embeddings of the device description can focus more on the words that contain more information about the classification of the device than less relevant words that provide little help to the semantic meanings of vector representations. Also, common words (stop words) are removed from the description before embedding. As shown in *Figure 1*, an empirical elbow rule is used and words like "of", "used", and "device" that appears in over one-fifth of the whole dataset is removed from the device description. After the sentence embeddings are derived, similarity scores are computed simply by deriving the cosine similarity between two sentence vectors.

The supervised FastText model that is used as a comparison in the experiments is composed of a simple linear model with rank constraint and a hierarchical SoftMax layer. This model utilizes n-gram features during its training to capture a certain amount of information about word order[14]. The model is trained with 60 epochs and a learning rate of 0.8. The input dimension is set to 300 in accordance with the pre-trained word embeddings.

**Sentence Transformer**

BERT is a deep bidirectional transformer-based pre-trained network that learned contextual relation and sentence dependency during the training[15]. Sentence embedding using BERT is usually done by taking the average of the last few output layers. However, it has been proven to underperform the average of Glove in sentence embedding due to the non-smooth anisotropic semantic space it introduces in the sentence embeddings[16]. Regarding this problem, Nils Reimer, et al[17], propose a new Siamese network based on BERT that uses a mean pooling layer by default that outputs a fixed 768-dimensional output vector and a cosine similarity computation above the Siamese network to derive the similarity score of two sentences. It achieved a state-of-the-art performance at that time and outperformed many other sentence encoding models like InferSent and Universal Sentence Encoder in most of the semantic textual similarity tasks. While the paper utilizes BERT as the base model, the same architecture can also be applied to other large pre-trained models such as RoBERTa and MPNet. These kinds of models are generally referred to as the sentence transformer. This paper explores the performance of several types of sentence transformers on the dataset.

Another advantage of sentence BERT is that it is computationally efficient. It computes sentence similarity with a much lower time complexity than BERT alone. Sentence BERT utilizes WordPiece as its tokenizer. WordPiece constitutes its library by merging most likely symbols from the training dataset and is not able to process OOV words. It represents them with an 'unknown' token instead. However, because the dataset in this paper contains a lot of unique and professional terms that are probably not included in the pre-trained vocabulary, this paper also tests the sentence transformer based on RoBERTa, which utilizes byte level BPE as its tokenizer[18]. Using bytes as the base vocabulary, byte-level BPE can handle any words by combining their constituting bytes. This paper also tests sentence transformer based on T5, a model developed by Google in 2020 that brought together the essence of many large pre-trained models and is pre-trained on C4 (Colossal Clean Crawled Corpus). T5 achieves state-of-the-art performance on many NLP benchmarks. T5 utilizes SentencePiece in conjunction with Unigram as its tokenizer[19]. While T5 is not able to consider OOV words, it considers spaces in building vocabulary. However, because T5 is trained on C4, a super large and clean corpus, there should not be a lot of OOV words. Lastly, this paper also tried sentence transformer on MPNet, a model developed by Kaitao Song, et al. MPNet combined the advantages of Masked Language Modeling (MLM) like BERT and Permuted Language Modeling like XLNet by incorporating extra positional information in the permuted-based loss function[20]. It is therefore able to understand a piece of text both based on its positional information and non-positional information. MPNet utilizes a tokenizer that is inherited from BERT. Considering that MPNet is also trained on a relatively large corpus(160GB), OOV words should not be a big problem too.

**GPT-3 Embeddings**

GPT-3 is a type of generative transformer that adopts almost the same structure as GPT-2 with stacks of transformer decoders except with attention layers of the Spare Transformer. It utilizes BPE as its tokenizer. BPE also can't handle OOV words and label them with 'unknown' tokens. It is trained with auto-regression and is well known for its superb contextual few-shot ability. Provided with just a few examples, the model is capable of generalizing on its own and delivering high performance on NLP tasks[21]. In this experiment, the Davinci semantic search and Davinci similarity embeddings from OpenAI are used which output 12288 dimensions of vectors for sentence representations. Cosine similarity is also applied to these embeddings to find the similarity score. This intuitively bears more resemblance to zero-shot learning.

|  | *average rank on correctly labeled device* | *average rank on wrongly labeled device* | *correlation with length\ significance* |
| --- | --- | --- | --- |
| *FastText* | 222.92 | 448.3 | *-0.15185 \0.47* |
| *GTP-3 Semantic Search* | 9.44 | 506.3 | *-0.19852 \0.34* |
| *GTP-3 Similarity* | 67.32 | 306.7 | *-0.06744 \0.75* |
| *Sentence BERT* | 181.92 | 809.3 | *-0.05002 \0.81* |
| *Sentence T5* | 34.16 | 793 | *-0.20034 \0.34* |
| *Sentence RoBERTa* | 162.28 | 292 | *-0.12397 \0.55* |
| *Sentence MPNet* | 14.04 | 296.3 | *0.18009 \0.40* |

*Table 2: various experiment results. Ranks are found by sorting cosine similarity of all possible labels in the CFR classification panels. Pearson correlation values are found between the word counts of the description and the ranks of the description. Significance is computed using a double-sided t-test.*

## IV. ANALYSIS

This experiment uses these methods by first finding the similarity ranks of the original labels of all correctly labeled devices. These methods are also evaluated on performance when the target devices are not correctly labeled. Finally, correlations are found between the lengths of the correctly labeled description text and the accuracy of all methods to explore the potential confounding factors in this experiment. The results are shown in *Table 2*.

As shown in *Figure 2*, which is the distribution of all similarity ranks predicted by the respective methods, FastText pretrained embedding, out of all methods, has the lowest performance with a large average rank of 222.92 and large variance with several outliers reaching ranks 1200, the random guess baseline. This is probably because sentence embedding is derived by taking the tf-idf average of the pre-trained embedding. Some descriptions in the CFR tend to be very comprehensive and describe all variants of a device type. This can be troublesome during tf-idf averaging because tf-idf might favor some of these variants that appear less often in the corpus. Another problem with averaging FastText pre-trained embedding is that it has no clues about the positions of the words in the sentences, which is especially tricky because the descriptions of medical devices include a lot of important verb-noun pairs that describe the utilities of the devices like "drain urine" and "clean vasculature", which will induce great amounts of information lost by mixing them up.

All other methods besides FastText embedding involve a self-attention mechanism and encode the positional information except that GPT is based on transformer decoder and BERT is based on transformer encoder and T5 is based on encoder-decoder. GPT-3 similarity also demonstrates a low performance when the CFR device type description is too general and comprehensive. It makes sense because adding too much information will alter the semantic meaning of the sentences. This doesn't cause trouble to GPT-3 semantic search embedding because it utilizes different embeddings for keys and queries. With the CFR descriptions being the queries and the target description being the key, it acts more like a keyword searching on the sentence level and therefore has higher endurance for the general and extra information in the queries.

While BERT is trained on sentence paired and employed next sentence prediction during its training, sentence BERT is also trained on the SNLI dataset, which contains 570,000 sentence pairs and the logical relations between them like entailment, contradiction, and neutral. Therefore, the sentence BERT similarity approach also captures the logical relations between the two sentences and is less sensitive to the extra information in the queries. However, unlike models that are based on permuted language models and able to encode stronger contextual information, BERT is a masked language model and less powerful as compared to models that encode both contextual and positional information like MPNet.

Sentence transformer like MPNet achieves much more accurate results on the dataset as expected. Sentence transformer based on T5 also achieves accurate results. Although T5 is also a masked language model, it is trained with a masked span infilling objective instead of masked token prediction like BERT. It, therefore, encodes more positional information by predicting a set of phrases instead of words. T5 is also trained on a much larger and cleaner dataset than any other pre-trained models. It also has a larger scale than most of the pre-trained models. Both MPNet and T5 achieve desirable results on the dataset with an average rank of 14.04 and 34.16 respectively.

We also trained the CFR dataset on the FastText supervised model with FastText pre-trained embeddings with 50 epochs and a learning rate of 0.8. We then evaluate the model on the 510(k) dataset. The model only achieves an accuracy of 0.25 on 100 most likely labels, which is much lower than finding the cosine similarity with FastText pre-trained embedding with tf-idf averaging, which achieves an accuracy of 0.64 using the same metrics. This makes sense because there is only one description for each label in the training set, and the model has a natural proclivity for overfitting. Therefore, a supervised model is not a reasonable choice in our case with such a limited dataset.

In conclusion, the performance on predicting correct labels, GPT-3 semantic search, MPNet, and T5 all perform accurately on the dataset, the correct labels are usually within the top 20 most likely labels.

Then this paper will give an examination of the incorrectly labeled device in the 510(k) dataset. Out of the three device descriptions that the author identified from the 510(k) dataset, one device is completely erroneously labeled, which means the original device label in the 510(k) document has no relation to the device description. In this case, as shown in *Figure 3*, all our methods have correctly assigned a very high rank for the device (device 18). The other two wrongly labeled devices are closely related to the labels they are assigned in word compositions except that they describe different items. For example, device 9 is a catheter set used to drain excess body fluids from the urinary to the abscess. However, it is classified as a biliary catheter that specializes in draining the biliary tract. While both descriptions contain keywords like body fluid, draining, and body tissues, all the methods fail in assigning a rank that is significantly larger than other correctly labeled devices. This probably provides a potential way of improving the model by incorporating the subject information in the embeddings (e.g., catheter set in this case).

Finally, this paper examines the effect of the lengths of the sentences being embedded on the accuracy of similarity comparison. Intuitively, methods like averaging word embeddings to form sentence embeddings are susceptible to the length of the sentence. Pearson coefficients are taken between the word counts of the description being classified and the rank of the prediction. Then a two-sided t-test is performed to determine the significance of the correlation. The test found no significant correlation between the length and the performance.

## V. DISCUSSION

Medical devices, according to [19], has many unique traits. It is constantly evolving; generations of devices are developed based on one another with more refined functionality. The mechanisms of these devices are often designed and well-understood as compared to pharmaceutical drugs, whose functional processes are sometimes complicated and cannot be clearly described. Several efforts have been made to improve the accuracy and efficiency of the FDA medical device regulation process. Starting from the 1990s, FDA adopted the Bayesian approach during the PMA examination. With the additional considerations of the prior information during trials and simulations, the FDA boosts its capability in making premarket approval decisions[22]. Other efforts like arguments by Terrie L. Reed, et al[23], have been made to help FDA establish a well-defined code to regulate the adverse event associated with medical devices. Donald L. Patrick, et al[24], also suggests using Patient-Reported Outcome to evaluate a medical device during clinical trials. All these methods help FDA to perform its duty of safeguarding the safety and effectiveness of medical devices.

|  | advantages | problems |
|---|---|---|
| Bayesian Statistics in trials | shorten the trial period and achieve the same accuracy with a smaller sample size with the incorporation of prior information | without enough detailed operation characteristics, it might take more simulations and increase the complexity of designing and planning out experiments in a Bayesian trial |
| Patient Reported Outcomes | measure the effectiveness of medical therapy in a more sensitive and specific way. | The poorly designed instrument might lead to misleading results |
| Adverse Event Problem Code | maintain and control events in a controlled manner. Less ad-hoc reports and issues | inefficient to retrieve old records of the adverse event report using the new code |

*Table 3: Several different methods are suggested to improve the accuracy/efficiency of the FDA approval process.*

However, to my best effort, the author found no related research being done so far on promoting the accuracy and efficiency of device classification. According to the official document of the 513 (g) request from the FDA, applicants need to manually identify the regulation number of their devices from the CFR classification panel, which contains over 2000 types of devices. If an applicant is unsure about the type of the devices, FDA also provides assistance through 513(g) requests, which took at most 60 days to provide an expert opinion on the type of the devices and their corresponding classifications.

With the new model that is built in this paper, applicants can simply type in brief descriptions of the types and functionalities of their devices, and the model will automatically display 10-15 most similar device types which, according to the performance of the sentence BERT and GPT-3 semantic search embedding, will contain the correct label with a very high chance. So instead of having to identify their device classifications from 2000 device types, applicants now only need to look through at most 20 types of devices to determine the classification of their device and which premarketing pathway they should adopt.

FDA can also use this model to identify completely irrelevant labeling in the submitted premarket applications. Although the approaches shown in this paper have shown the incapability of identifying incorrect labeling that is close in word compositions to the correct labeling, all the methods show high accuracy in identifying labels that are completely unrelated to the device descriptions.

While FDA receives over 18000 premarket applications annually and is still increasing, automation or partial automation of the premarket regulation becomes an inevitable process with incredible potential. Introducing algorithms in the device classification saves time not only for applicants but also for FDA to verify the applicant-classified results. This increase in efficiency can either be utilized to dilute the workload of the FDA or allow FDA to focus more on the PMA clinical trials process, which is much more imperative.

Several ways might help improve the model's accuracy, and they need further experimentation. Firstly, a more extensive dataset might help build a supervised model specializing in device classifications. With a larger dataset, we can train models based on existing large pre-trained models like ERNIE and XLNet, which will be able to capture the semantic and contextual meaning of the sentences better.

Secondly, as mentioned in the analysis section, it is possible to incorporate the subject's information in the embeddings. However, this would probably need to train a small network to extract the subject information first. Then, inspired by the idea of position embedding, this information can be incorporated into the classification model by adding an extra embedding layer that encodes the subject of the sentence, which is especially important in device classification.

One another problem to consider is that the sentences we match, the CFR description of the device types, and the description of a specific device that needs to be classified are not strictly the same in a logical sense. The CFR descriptions provide much more general information on classes, while the device descriptions are specific cases under these classes. These two kinds of texts should not possess thorough similarity inherently. Therefore, it may be possible to encompass logical information, like sentence BERT, for the model to better understand the links between keys and queries. Further research can be done on how these large language models transfer this logical relationship behind sentences.

## VI. CONCLUSION

This paper proposes an NLP method that assistances the FDA regulates its premarket devices. Several large pre-trained models are directly tested on the dataset by providing a reference dataset (the CFR description). If suitable models like MPNet and T5 are used, the classification results are still reasonably accurate (average rank at around 10) even without training data. This probably also reveals the capability of these large language models on other resource-limited tasks. While multimodal learning models nowadays, for example, BEIT-3 and CLIP, are performing exceedingly well in popular objectives like image production and language generation, which own an enormous amount of training data, it also sheds light on improving the performance of low-resource tasks. While these large pre-trained models provide "understanding" in language and images, these low-resource tasks can be the downstream task that offers little or no training data but still achieve good results.

On the other hand, as these large models are becoming more and more widely used, researchers are beginning to perform analysis on how fine-tuning and bias might impede the transfer of knowledge to other downstream tasks.[25] we must realize that often when we are transferring knowledge from these large language models to downstream tasks, we are testing in an environment with distinct data distribution than the training environment. Ananya Kumar, et al have shown that fine-tuning all model parameters on downstream tasks can be potentially harmful to the out-of-distribution accuracy[26], which implies the urgency of careful analysis of the performance of models suffering from distribution shifting. Relevant tests like [27] [28] [29] are needed to provide more information about the reaction of the model performance under different correction approaches. As new information is continuously emerging, and the models are destined to adapt accordingly, more efficient methods are needed for NLP models to shift to the target domain while maintaining the knowledge it gains from the older one.

However, a more extensive dataset also brings the problem of biasing. Larger datasets are harder to supervise. As most of the training data for these large models are crawled directly from the internet, there might be potential inaccuracy or inaccuracy in these training data, leading to bias in the models and the results that are hard to measure.

This problem can inconspicuously become a significant risk, especially in policymaking, like regulating premarket devices. If flawed products are allowed to flow into the market or biased policies and decisions are made by these algorithms, the negative consequence is immeasurable. Therefore, more reliable verification/secondary methods are crucial in this social application. Methods like data mining and anomaly/fraud detection should be developed to supervise training data and predictions. Computational methods and algorithms in anomaly detection like [30] and graph anomaly detection should be brought to the field of text data to recognize better bias in these large-scale corpora.

While the training data alone can be unfair and biased, it can also lead to unstable models prone to inaccurate data fluctuations like adversarial perturbations. This fact is also shown through experiments in this paper when all models wrongly classified an all-purpose catheter as a biliary catheter. Research has shown that the vulnerability of these models can be attributed to no-robust features[31]. Many advanced approaches have been developed to handle this problem like [32], [33], [34], all devoted to eliminating the learned unstable features from the models. More research on adversarial attacks is needed to help us understand how these large language models understand and process texts.

However, this will lead to further ethnic problems by using algorithms to regulate algorithms. More discussions need to be made on the topic, like the one done by Chesterman[7] and Sunstein[8] are needed. More experiments are also required to test this kind of regulation's potential risk and applicability.

As neural networks and other machine learning models are already widely used in economics, law, and other human-centered studies, better usage of these models and data can benefit society immeasurably in the near future.

## VII. SUPPLEMENT

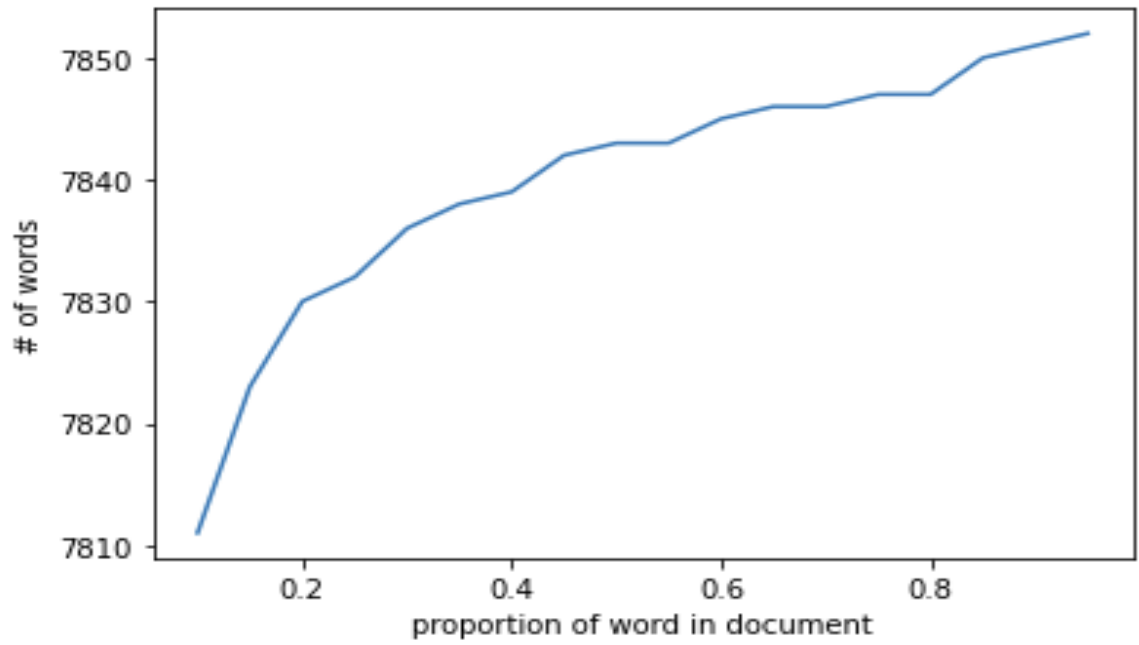

*Figure 1: number of words remaining after stop words are removed versus the proportion of words appearing in the entire dataset to be considered stop words. An elbow rule is adopted, and words appearing in more than 1/5 of the descriptions in the dataset are removed.*

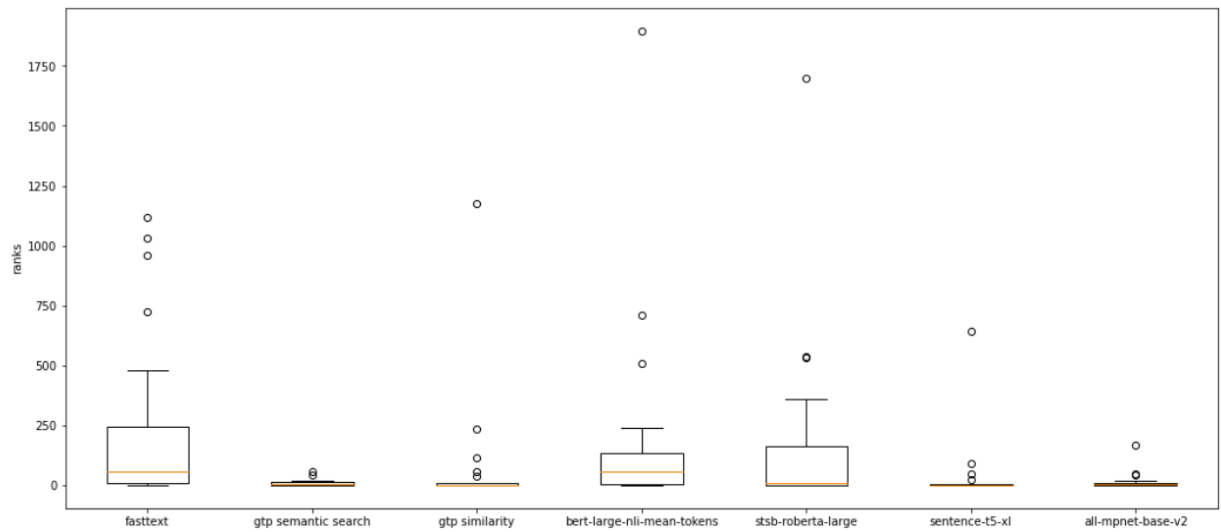

*Figure 2: Distribution of prediction ranks using different methods.*

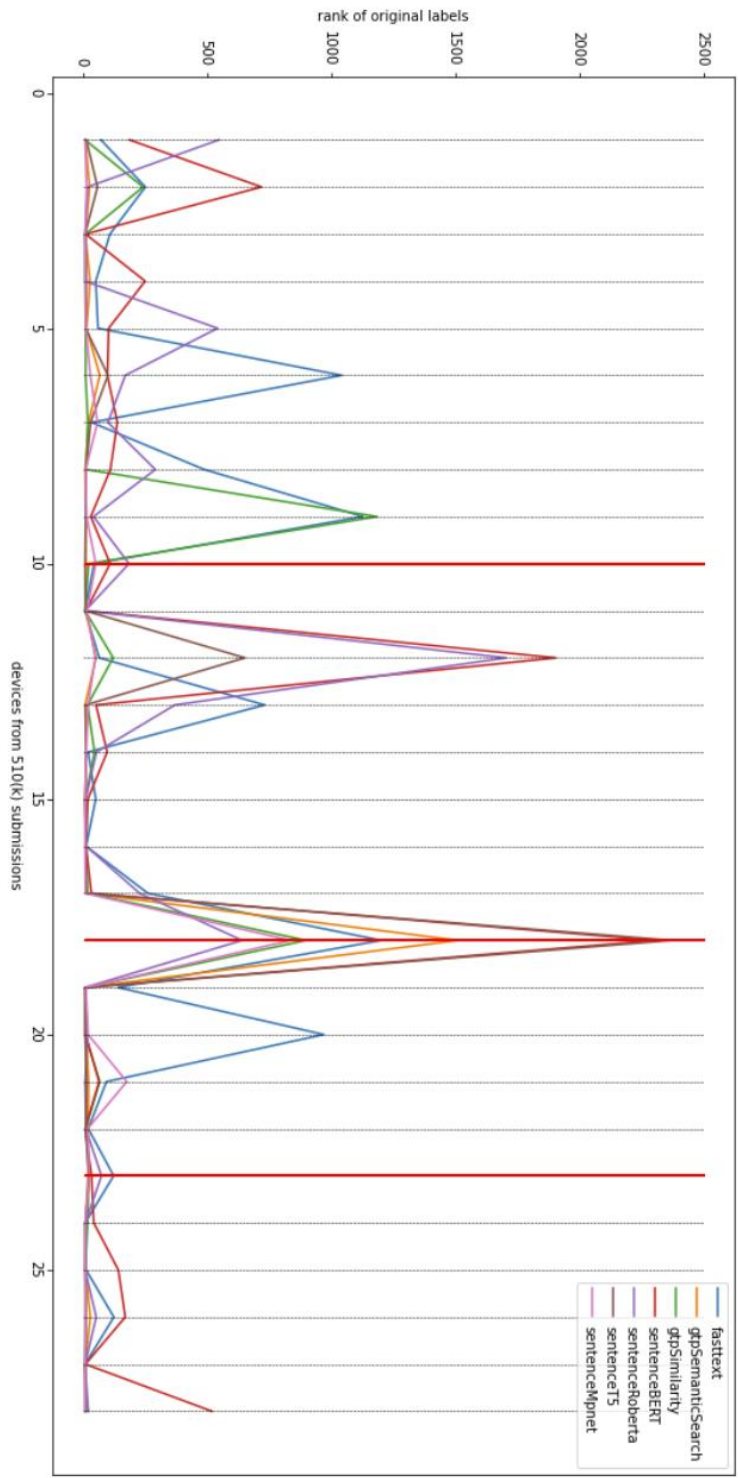

Figure 3: Different ranks are predicted by different methods across the 510(k) dataset. Wrongly labeled data are marked with red vertical lines.